\documentclass[pdflatex,sn-basic,Numbered,iicol]{sn-jnl}

\usepackage{graphicx}%
\usepackage{multirow}%
\usepackage{amsmath,amssymb,amsfonts}%
\usepackage{booktabs}%
\usepackage{xcolor}%
\usepackage{manyfoot}%
\usepackage{enumitem}%
\usepackage{wrapfig}%
\usepackage{rotating}%
\usepackage{bookmark}%
\renewcommand{\orcid}[1]{}%

\newcommand{\vR}{\mathbf{R}}
\newcommand{\vI}{\mathbf{I}}
\newcommand{\vN}{\mathbf{N}}
\newcommand{\vP}{\mathbf{P}}
\newcommand{\vD}{\mathbf{D}}
\newcommand{\vL}{\mathbf{L}}

\newcommand{\etal}{\textit{et~al.}}

\begin{document}

\title[Scene-Adaptive Tone Curves for Low-Light 3DGS]{Scene-Adaptive Nonlinear Tone Curves for Pseudo Ground-Truth Generation in Low-Light 3D Gaussian Splatting}

\author[1,2]{\fnm{Mingzhe} \sur{Lyu}}\orcid{0009-0007-3551-8909}

\author[2]{\fnm{Jinqiang} \sur{Cui}}\orcid{0000-0002-7833-1876}

\author*[1]{\fnm{Hong} \sur{Zhang}}\email{hzhang@sustech.edu.cn}\orcid{0000-0002-1677-6132}

\affil[1]{\orgdiv{Shenzhen Key Laboratory of Robotics and Computer Vision}, \orgname{Southern University of Science and Technology}, \orgaddress{\city{Shenzhen}, \postcode{518055}, \country{China}}}

\affil[2]{\orgname{Pengcheng Laboratory}, \orgaddress{\city{Shenzhen}, \postcode{518108}, \country{China}}}

\abstract{Low-light novel view synthesis is challenging because dark multi-view images contain noise, weak structural detail, and compressed dynamic range. Recent 3D Gaussian Splatting (3DGS) methods address these challenges by generating pseudo ground-truth (pseudo-GT) images as supervision targets when paired normal-light references are unavailable. Existing pseudo-GT methods apply a uniform linear gain to all pixels, which clips bright regions while providing insufficient enhancement in dark regions, limiting reconstruction quality. We observe that nonlinear tone mappings, long established in 2D low-light enhancement, have not been explored for pseudo-GT generation in 3D reconstruction. Accordingly, we propose a scene-adaptive nonlinear tone-curve framework that replaces linear pseudo-GT with nonlinear alternatives. The framework introduces percentile-based normalisation for scene-agnostic curve application, a scene-adaptive offset for automatic black-level adjustment, and two complementary curves: Adaptive SoftExp (ASE), a bounded exponential curve, and Adaptive Poly3 (AP3), a data-driven cubic polynomial. The module changes only the pseudo-GT computation and leaves the 3DGS backbone unchanged. Experiments on three benchmarks covering 21 scenes show that both curves consistently outperform the linear baseline with PSNR improvements up to $+$4.34\,dB on LOM and $+$3.25\,dB on RealX3D. Both curves achieve similar performance despite their different mathematical forms, suggesting the improvement is curve-agnostic. Code is available at \url{https://github.com/lvmingzhe/adaptiveToneCurve}.}


\maketitle

\section{Introduction}
\label{sec:intro}
Novel view synthesis (NVS) renders photorealistic images from novel viewpoints given multi-view captures, with broad applications in augmented/virtual reality, autonomous driving, and digital content creation. Neural Radiance Fields (NeRF)~\citep{mildenhall2020nerf} and 3D Gaussian Splatting (3DGS)~\citep{kerbl20233dgs} have enabled substantial progress under well-lit conditions, yet real-world captures are frequently acquired in low-light environments such as nighttime surveillance, underground exploration, and poorly lit interiors, where images exhibit low signal-to-noise ratio, weak structural detail, and compressed dynamic range. These degradations jointly impair both geometry estimation and appearance optimisation, making standard NVS pipelines produce poor reconstructions.

Several approaches have been proposed for low-light NVS, focusing primarily on network architecture and scene representation. Early attempts apply 2D enhancement methods such as Zero-DCE~\citep{guo2020zerodce} and SCI~\citep{ma2022sci} as pre- or post-processing around standard NVS pipelines, but per-view enhancement does not enforce multi-view consistency. End-to-end NeRF-based methods, including Aleth-NeRF~\citep{cui2024alethnerf} and I$^2$-NeRF~\citep{liu2025i2nerf}, achieve improved quality by integrating illumination modelling into the radiance field at the cost of long training times. LITA-GS~\citep{zhou2025litags} achieves state-of-the-art quality at real-time rendering speed with illumination-invariant structure priors. These methods have advanced the field through representation design, loss formulation, and dedicated modules. However, none has systematically examined the pseudo-GT generation that serves as the primary supervision signal when paired normal-light references are unavailable.

In LITA-GS, pseudo-GT images are generated by multiplying each low-light frame by a uniform gain factor and clamping the result to the valid intensity range. This linear strategy has two inherent limitations. First, the uniform gain amplifies all pixels equally: a factor large enough to lift dark regions into a visible range inevitably pushes brighter pixels beyond the representable range, where they are clipped and lose gradient signal. Second, the linear mapping provides no intensity-adaptive control, treating shadow and highlight regions identically. Notably, in 2D low-light enhancement, nonlinear tone mappings have long been the standard for bridging the gap between dark and bright images~\citep{guo2020zerodce,ma2022sci}, but this principle has not been applied to pseudo-GT generation in 3D reconstruction.

\begin{figure*}[!t]
\centering
\includegraphics[width=0.95\textwidth]{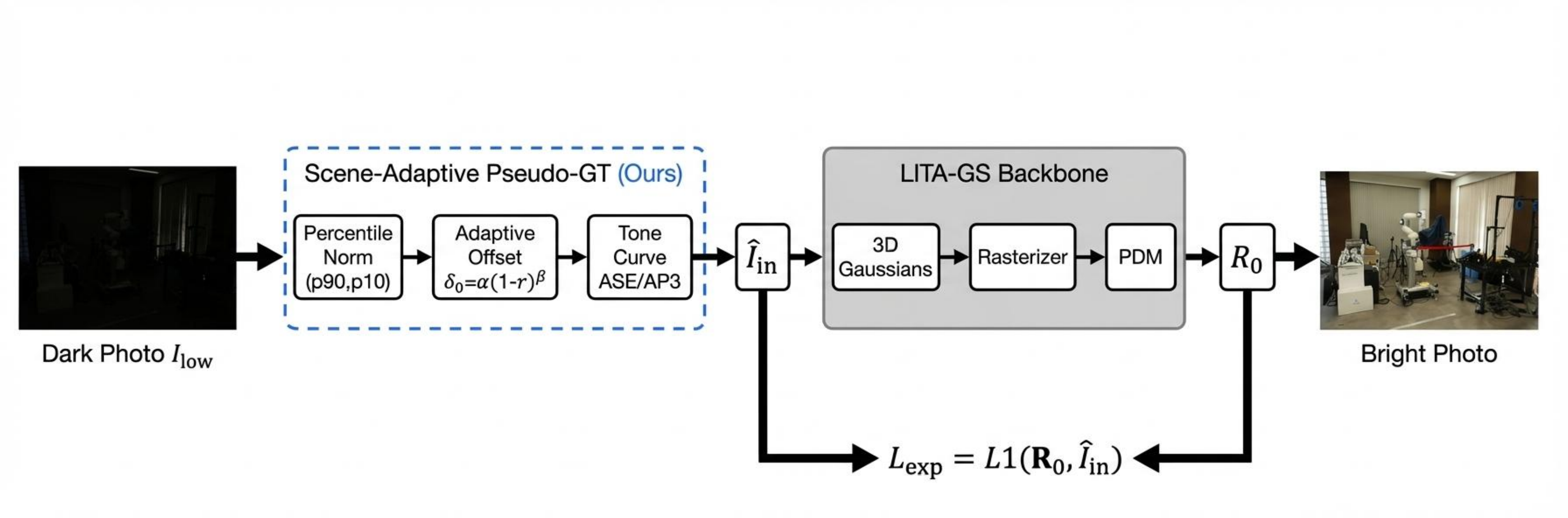}
\vspace{-0.6em}
\caption{System overview. Our scene-adaptive nonlinear tone-curve module (dashed box) replaces the original linear pseudo-GT generation in LITA-GS. The module computes percentile-based normalisation, an adaptive offset $\delta_0 = \alpha(1-r)^\beta$, and applies either the ASE or AP3 tone curve to produce the pseudo-GT $\hat{\vI}_{\mathrm{in}}$. The LITA-GS backbone (3D Gaussians, rasteriser, PDM) remains entirely unchanged. The exposure control loss $\mathcal{L}_{\mathrm{exp}} = \mathcal{L}_1(\vR_0, \hat{\vI}_{\mathrm{in}})$ supervises the rendered RGB output.}
\label{fig:architecture}
\end{figure*}

To address these limitations, we propose a scene-adaptive nonlinear tone-curve framework for pseudo-GT generation in low-light 3DGS (Fig.~\ref{fig:architecture}). A nonlinear tone curve inherently applies stronger lifting in dark regions and gentler compression in bright regions, addressing both limitations of linear scaling without requiring architectural changes. The framework consists of three components: percentile-based normalisation that makes the curve shape scene-agnostic regardless of absolute exposure, a scene-adaptive offset that adjusts the black-level lift per image based on robust image statistics, and two complementary tone curves, Adaptive SoftExp (ASE) and Adaptive Poly3 (AP3), that instantiate the nonlinear mapping from different design principles. The entire module is a drop-in replacement for the linear pseudo-GT, leaving the LITA-GS backbone unchanged and adding less than 0.5\,ms of overhead per training iteration.

Experiments on three benchmarks covering 21 scenes validate the proposed framework. On LOM~\citep{cui2024alethnerf} (5 scenes, 5K iterations), both ASE and AP3 improve mean PSNR by over 4\,dB compared to the linear baseline. On RealX3D~\citep{liu2025realx3d} (9 scenes, 15K iterations), AP3 improves NVS PSNR by $+$3.25\,dB and ASE by $+$2.57\,dB. On MipNeRF360-varying~\citep{barron2022mipnerf360} (7 scenes, 15K iterations), both curves improve over the baseline on all three mean metrics (PSNR, SSIM, LPIPS). Notably, ASE and AP3 achieve similar performance across all benchmarks despite their different mathematical forms, confirming that the improvement stems from the nonlinear mapping itself rather than any specific curve parametrisation. The framework uses a unified offset configuration across all three benchmarks, requiring only a dataset-level brightness gain for adaptation.

Our contributions are summarised as follows:
\begin{enumerate}[leftmargin=*]
\item We identify the pseudo-GT tone curve as an overlooked design dimension in low-light 3DGS and show that nonlinear curves consistently outperform linear scaling on PSNR, SSIM, and LPIPS across three benchmarks covering 21 scenes. Two structurally different curves achieve similar gains, providing evidence that the nonlinear mapping itself, rather than any specific parametrisation, is the primary driver of quality gains.

\item We instantiate the framework with two complementary curves: Adaptive SoftExp (ASE), a physically grounded exponential curve with natural boundedness, and Adaptive Poly3 (AP3), a data-driven cubic polynomial selected from a systematic search over 22 candidate families. Both are drop-in replacements that modify only the pseudo-GT computation and leave the 3DGS backbone unchanged.

\item We introduce a scene-adaptive offset based on percentile statistics that automatically adjusts the black-level lift per image. Combined with a dataset-level brightness gain, this mechanism enables a unified configuration to generalise across all three benchmarks without per-scene hyperparameter tuning.
\end{enumerate}

\section{Related Work}
\label{sec:related}

\subsection{Low-Light Novel View Synthesis}

NeRF~\citep{mildenhall2020nerf} and 3DGS~\citep{kerbl20233dgs} achieve photorealistic novel view synthesis but degrade severely when the inputs are low-light images. Earlier radiance-field work addresses high dynamic range (HDR) or noisy raw capture by incorporating exposure and tone modelling into volumetric rendering~\citep{mildenhall2022rawnerf,huang2022hdrnerf}. Recent HDR-oriented 3DGS work further extends this direction to video reconstruction~\citep{chen2025evhdrgs}. These methods focus on HDR or raw reconstruction rather than supervisory tone-curve design. Recent low-light NVS methods can be grouped by where in the pipeline they primarily address the brightness deficiency; individual methods may combine techniques from multiple stages.

\textit{(i) Input-side:} These methods modify the training images before 3D reconstruction. LLGS~\citep{wang2025llgs} preprocesses with gamma correction; Cao~\etal~\citep{cao2025semanticgs} apply semantic guidance; L$^2$DGS~\citep{kumar2026l2dgs} extends input-side supervision to dynamic scenes via a learned well-lit-to-low-light transformation. LITA-GS~\citep{zhou2025litags} also generates pseudo-GT by linearly scaling low-light frames, though its primary contributions lie in rendering-side modules.

\textit{(ii) Representation-side:} These methods encode illumination explicitly within the 3D scene model. Aleth-NeRF~\citep{cui2024alethnerf} learns a transmittance-attenuating concealing field; I$^2$-NeRF~\citep{liu2025i2nerf} models low-light scenes as a Beer--Lambert absorption medium with a Bright-Channel transmittance prior; LLNeRF~\citep{wang2023llnerf} and Bright-NeRF~\citep{wang2025brightnerf} decompose illumination, noise, and colour from sRGB and raw inputs in an unsupervised manner, respectively; GloNeRF~\citep{liu2025glonerf} boosts multi-view consistency under low light.

\textit{(iii) Rendering-side:} These methods embed enhancement in the rendering pipeline. LITA-GS~\citep{zhou2025litags} introduces illumination-invariant structure priors and a progressive denoiser; Luminance-GS~\citep{cui2025luminancegs} and Gaussian-in-the-Dark~\citep{ye2024gaussiandark} embed view-adaptive curves and sensor-response correction inside Gaussian rendering; DarkGS~\citep{zhang2024darkgs} learns a neural illumination model for relighting dark scenes; and LuSh-NeRF~\citep{qu2024lushnerf} jointly models low-light noise and camera-shake blur for NeRF. A concurrent work, MERID-GS~\citep{yin2026lightup}, decouples illumination and reflectance via explicit Retinex decomposition for few-shot 360$^\circ$ synthesis.

Despite this growing diversity, input-side methods uniformly rely on simple brightness transforms such as linear scaling or gamma correction, and no prior work systematically examines whether replacing these with a nonlinear tone curve can improve reconstruction quality. Our work fills this gap.

\subsection{Low-Light Image Enhancement}

2D low-light enhancement provides the conceptual foundation for our pseudo-GT design and is typically organised by enhancement paradigm.
\textit{Retinex-based decomposition} traces a progression from RetinexNet's paired supervision~\citep{wei2018retinexnet}, through KinD/KinD++~\citep{zhang2019kind,zhang2021beyond} and the unfolded URetinex-Net~\citep{wu2022uretinex}, to the one-stage transformer Retinexformer~\citep{cai2023retinexformer} and Retinex-memory mechanisms~\citep{zhang2026retinexmemory}.
\textit{Curve-based} approaches learn higher-order tone curves~\citep{guo2020zerodce,li2022zerodcepp}, self-calibrated illumination maps~\citep{ma2022sci}, or semantic-guided brightness curves~\citep{wang2025sbcnet}.
\textit{End-to-end} pipelines range from decomposition--restoration designs~\citep{pan2024dicnet,yu2023twostage} and the MIRNet multi-scale backbone~\citep{zamir2020mirnet}, to lightweight or UHD transformers~\citep{cui2022iat,wang2023llformer}, unpaired adversarial supervision~\citep{jiang2021enlightengan}, SNR-aware enhancement~\citep{xu2022snraware}, and recent methods exploring illumination magnitude control, natural-illumination estimation, invertible networks, and multi-scale fusion~\citep{shi2024maco,singh2024nplie,zhang2024invertible,dong2025dednet}.
\textit{Diffusion-based} methods recently push enhancement quality further by coupling Retinex priors with generative refinement~\citep{jiang2024lightendiffusion,yi2025diffretinexpp,liu2026compbalanced}.
A common finding across these paradigms is that \emph{nonlinear mappings} are essential for bridging the distributional gap between dark and bright images. Our work transfers this principle from 2D enhancement to pseudo-GT generation in the 3DGS training loop, showing that the choice of nonlinear curve shape broadly improves the quality of 3D reconstruction.

\subsection{Tone Mapping}

Tone mapping has a long history in display and rendering, where the goal is to compress HDR content into a representable range. Our work draws on three lines from this literature.

\textbf{Global tone mapping operators}~\citep{reinhard2002photographic,drago2003adaptive} apply a single nonlinear curve (logarithmic, sigmoid, or filmic) across the entire image, with Hable's filmic curve~\citep{hable2010uncharted} from real-time game rendering following the same paradigm. These operators assume the input already captures the full scene dynamic range; their nonlinear shape governs how highlights are compressed but does not address shadow lifting on heavily underexposed low-light inputs.

\textbf{Local and spatially-adaptive tone mapping}~\citep{durand2002bilateral,mantiuk2008display} processes different image regions with different curves based on local luminance statistics, capturing the intuition that a single global curve cannot fit all spatial regions. This motivates our scene-adaptive offset that adjusts the black-level lift per image based on robust image-level percentile statistics.

\textbf{Inverse tone mapping}~\citep{banterle2017advanced} addresses the reverse direction, reconstructing HDR content from low dynamic range inputs, and is most closely related conceptually to our pseudo-GT generation, where we map dark inputs toward bright targets. However, prior inverse tone mapping methods focus on display-side reconstruction and do not consider the supervisory role of these curves in 3D scene reconstruction.

In summary, classical tone mapping provides the nonlinear curve vocabulary we adopt (polynomial and exponential families), but these operators were designed for display-side compression, not for generating supervision targets in 3D reconstruction. Our work repurposes this vocabulary for pseudo-GT generation and demonstrates that the nonlinear mapping itself is a key factor in reconstruction quality. The framework is orthogonal to representation-level 3DGS advances such as Mip-Splatting~\citep{yu2024mipsplatting}, Scaffold-GS~\citep{lu2024scaffoldgs}, Deformable 3D Gaussians~\citep{yang2024deformablegs}, and 2D Gaussian Splatting~\citep{huang2024twodgs}.

\section{Method}
\label{sec:method}

We propose a scene-adaptive nonlinear tone-curve framework that replaces the linear pseudo-GT in LITA-GS~\citep{zhou2025litags}. As illustrated in Fig.~\ref{fig:architecture}, the framework consists of three components: (1)~percentile-based normalisation for scene-agnostic curve application, (2)~a scene-adaptive offset that adjusts the black-level lift, i.e.\ the minimum output brightness assigned to dark pixels, per image, and (3)~two complementary tone curves (ASE and AP3) that instantiate the nonlinear mapping. Our module serves as a \emph{drop-in replacement} for the original linear pseudo-GT and leaves the LITA-GS~\citep{zhou2025litags} backbone unchanged throughout the entire training process.

\subsection{Preliminaries: LITA-GS Pipeline}
\label{sec:prelim}

LITA-GS~\citep{zhou2025litags} augments each 3D Gaussian primitive with four additional attribute channels: (1)~an illumination-invariant \textbf{structure prior} extracted via colour-invariant convolutions (CIConv2d) based on Kubelka--Munk reflectance theory, (2)~a \textbf{depth} attribute supervised by monocular depth estimates, (3)~an \textbf{illumination feature} that decomposes the scene into light-dependent and light-independent components, and (4)~a \textbf{noise} attribute modelling sensor noise in dark images.

During rasterisation, these attributes produce five parallel maps: the enhanced RGB image $\vR_0$, structure prior $\vP_r$, depth $\vD_r$, illumination $\vL_r$, and noise $\vN_{GS}$. A \textbf{Progressive Denoising Module (PDM)} with $K\!=\!3$ cascaded stages iteratively refines the rendered image: $\vR_{k+1} = \vR_0 - \vN_{k+1}$. A learnable tonemapper reconstructs the low-light observation as $\hat{\vI}_{\mathrm{low}} = \vR_K \odot \vL_r$. The total loss combines exposure control ($\mathcal{L}_{\mathrm{exp}}$), structure consistency ($\mathcal{L}_{\mathrm{str}}$), denoising ($\mathcal{L}_{\mathrm{de}}$), and low-light reconstruction ($\mathcal{L}_{\mathrm{rec}}$).

Among these losses, the exposure control loss $\mathcal{L}_{\mathrm{exp}} = \mathcal{L}_1(\vR_0, \hat{\vI}_{\mathrm{in}})$ provides the dominant gradient signal to the Gaussian colour coefficients, where the pseudo ground-truth $\hat{\vI}_{\mathrm{in}}$ is the supervision target. In the original LITA-GS~\citep{zhou2025litags} framework, this pseudo-GT is generated by linearly scaling each low-light frame:
\begin{equation}
  \hat{\vI}_{\mathrm{in}} = \mathrm{clamp}\!\left(\frac{\theta}{\mathrm{mean}(\vI_{\mathrm{low}})} \cdot \vI_{\mathrm{low}},\; 0,\; 1\right),
  \label{eq:linear_gt}
\end{equation}
where $\theta$ is a fixed exposure hyperparameter that controls the target brightness level. The quality of $\hat{\vI}_{\mathrm{in}}$ therefore directly influences the upper bound on reconstruction quality. As discussed in Section~\ref{sec:intro}, this linear strategy clips highlights and provides no intensity-adaptive control; we introduce nonlinear replacements as detailed in the following subsections.

\subsection{Percentile Normalisation}
\label{sec:normalisation}

The mean-based normalisation in Eq.~\ref{eq:linear_gt} is sensitive to outlier pixels and produces unstable scaling across views with different exposure levels. We replace it with percentile-based normalisation for robustness. Given a low-light input $\vI_{\mathrm{low}} \in [0,1]^{3 \times H \times W}$, we first compute its luminance channel using the ITU-R BT.601 standard:
\begin{equation}
  Y = 0.299R + 0.587G + 0.114B
\end{equation}
and extract two robust order statistics:
\begin{equation}
  p_{90} = \mathrm{quantile}(Y, 0.90), \quad p_{10} = \mathrm{quantile}(Y, 0.10).
\end{equation}
The normalised input $t = \vI_{\mathrm{low}} / p_{90}$ maps the bulk of pixels into $[0, 1]$ regardless of absolute exposure level, making the subsequent curve shape \emph{scene-agnostic}. Unlike mean-based normalisation (Eq.~\ref{eq:linear_gt}), percentiles are robust to outliers and provide stable normalisation across different views.

\subsection{Scene-Adaptive Offset}
\label{sec:offset}

The key scene descriptor is the \textbf{shadow compactness ratio}:
\begin{equation}
  r = \frac{p_{10}}{p_{90}} \in [0, 1].
  \label{eq:r}
\end{equation}
A large $r$ indicates less severe shadow compression; a small $r$ indicates severely compressed shadows with limited recoverable detail, which require stronger black-level lifting. For example, on the LOM~\citep{cui2024alethnerf} dataset, the \textit{bike} scene has $r\!=\!0.37$ (bright backgrounds preserve shadow detail) while \textit{shrub} has $r\!=\!0.12$ (dense foliage produces dark shadows with limited detail).

The adaptive offset $\delta_0$ is defined as:
\begin{equation}
  \delta_0 = \alpha \cdot (1 - r)^{\beta},
  \label{eq:offset}
\end{equation}
where $\alpha$ sets the maximum offset magnitude (upper bound when $r \to 0$) and $\beta = 2.5$ governs the decay rate. We fix $\alpha\!=\!0.12$ and $\beta\!=\!2.5$ across all three benchmarks; these values were obtained via grid search on the BlueHawaii scene and validated by the multi-scene ablation (Section~\ref{sec:ablation}), with no per-scene tuning required.

The per-scene offset values for the five LOM scenes range from $\delta_0\!=\!0.038$ on \textit{bike} ($r\!=\!0.37$, $-63\%$ vs.\ the fixed cubic intercept of 0.103, preventing over-brightening) to $\delta_0\!=\!0.087$ on \textit{shrub} ($r\!=\!0.12$, retained for shadow recovery). This adaptive mechanism avoids the failure mode of fixed offsets, which over-brighten well-lit scenes and under-brighten dark scenes (see the \textit{bike} case study in Section~\ref{sec:bike_analysis}). The complete per-scene offset table is provided in Appendix~\ref{sec:supp_offset_sensitivity}.

\subsection{Adaptive SoftExp (ASE)}
\label{sec:ase}

The soft-exponential function $1 - e^{-x}$ is a bounded concave operator widely used as a tone-mapping primitive: it monotonically maps $[0,\infty)$ to $[0,1)$, lifts shadows linearly near zero, and saturates highlights smoothly. We compose it with the adaptive offset:
\begin{equation}
  f_{\mathrm{ASE}}(x) = \delta_0 + (1 - \delta_0) \cdot (1 - e^{-g \cdot t})
  \label{eq:ase}
\end{equation}
where $t = x / p_{90}$, $\delta_0$ is defined in Eq.~\ref{eq:offset}, and $g$ is the gain parameter controlling curve steepness. Rather than fixing $g$, we make it \emph{brightness-adaptive}:
\begin{equation}
  g = g_0 + c \cdot \bar{Y},
  \label{eq:gadapt}
\end{equation}
where $\bar{Y}$ is the mean luminance of the input low-light image, $g_0\!=\!1.0$ is the base gain, and $c$ is a global scaling constant. Brighter low-light images (higher $\bar{Y}$) indicate less extreme underexposure, but the $p_{90}$ normalisation provides less stretch for these images and can leave the pseudo-GT too dim. The $c \cdot \bar{Y}$ term compensates by increasing the curve gain in proportion to absolute brightness. The constant $c$ is selected per dataset (not per scene); we use $c\!=\!3$ on LOM and $c\!=\!4$ on RealX3D~\citep{liu2025realx3d} for ASE, and $c\!=\!0$ on MipNeRF360-varying~\citep{barron2022mipnerf360}, where the training views already span a wide brightness range, making additional gain adaptation unnecessary.

\textbf{Properties.} (i)~\textit{Bounded concave shape}: $f_{\mathrm{ASE}}(0) = \delta_0$ (adaptive black level), $\lim_{x\to\infty} f_{\mathrm{ASE}}(x) = 1$ (natural saturation), and strictly concave on $t > 0$, jointly compressing highlights while lifting shadows. (ii)~\textit{Smoothness}: the exponential core is $C^\infty$, providing well-defined gradients everywhere without clamp discontinuities. (iii)~\textit{Physical interpretation}: $1 - e^{-gt}$ can be interpreted as a simple absorption-like response of a sensor with gain $g$, connecting the curve to well-understood imaging physics.

\subsection{Adaptive Poly3 (AP3)}
\label{sec:ap3}

ASE has a simple bounded form. A data-driven curve provides a complementary option that can better match the brightness statistics favoured by the L1+SSIM reconstruction objective. Through systematic evaluation of 22 parametric curve families on the BlueHawaii scene (full leaderboard in Appendix~\ref{sec:supp_curve_search}), a cubic polynomial achieved the best validation PSNR:
\begin{equation}
  p(t) = 0.072t^{3} - 0.438t^{2} + 0.995t,
  \label{eq:ap3_poly}
\end{equation}
where $t = x / p_{90}$ and $\delta_0 = \alpha(1-r)^\beta$ as before. The polynomial coefficients were fitted via 2D regression on paired low/GT images from the BlueHawaii scene and then \emph{frozen}: they are never updated during 3DGS training and are not tuned per scene. Like ASE, AP3 incorporates a brightness-adaptive gain:
\begin{equation}
  f_{\mathrm{AP3}}(x) = \mathrm{clamp}\!\left(\delta_0 + p(t)\,b,\;0,\;1\right),
  \label{eq:ap3_full}
\end{equation}
where $b = 1 + c \cdot \bar{Y}$ scales the polynomial output based on the mean luminance $\bar{Y}$, using the same constant $c$ as in Eq.~\ref{eq:gadapt}. Note that $c$ enters at different positions in the two curves: inside the exponential for ASE ($g\!=\!g_0\!+\!c\bar{Y}$) and outside the polynomial for AP3 ($b\!=\!1\!+\!c\bar{Y}$), reflecting each curve's natural parameterisation. For brighter low-light images, $b > 1$ steepens the curve to compensate for the reduced stretch from the $p_{90}$ percentile normalisation step.

\textbf{Properties.} (i)~\textit{Near-linear initial slope}: the coefficient of $t$ is $0.995 \approx 1$, providing strong, nearly proportional lifting in the shadow-to-midtone range. (ii)~\textit{Cubic rolloff}: the negative quadratic term ($-0.438t^2$) compresses highlights for $t > 1$, preventing over-saturation. (iii)~\textit{Black-level lift}: $\delta_0$ raises the entire curve, recovering detail where shadow compression is most severe.

\textbf{Trade-off.} ASE is naturally bounded. The AP3 polynomial is unbounded and may produce values outside $[0, 1]$ for extreme $t$; a final clamp is therefore applied, though it affects fewer than 1\% of pixels across all scenes.

\subsection{Integration into LITA-GS}
\label{sec:integration}

\begin{figure}[t]
\centering
\includegraphics[width=\linewidth]{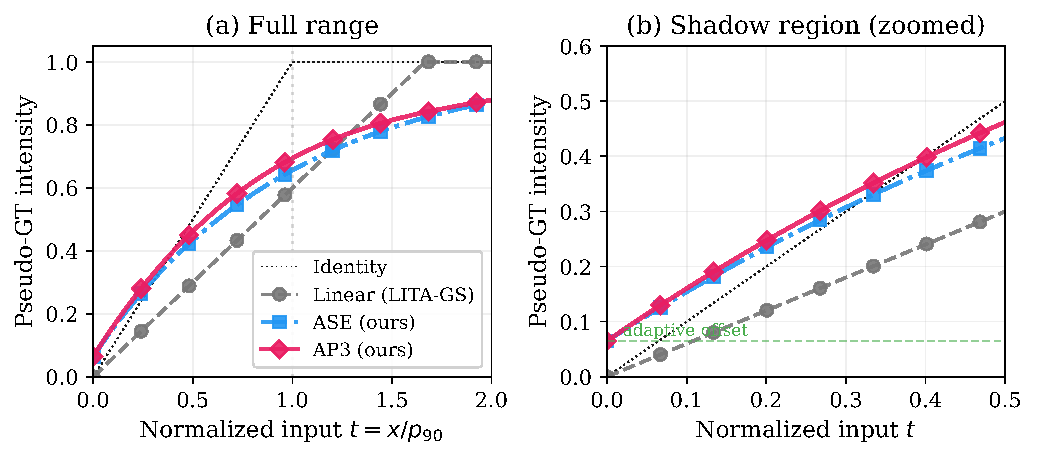}
\caption{Tone curve comparison (legend in panel). ASE and AP3 both provide nonlinear enhancement, while linear scaling clips at 1.0. The adaptive offset $\delta_0$ lifts the black level, with AP3's near-linear slope providing stronger midtone lifting than ASE's concave shape.}
\label{fig:curves}
\end{figure}

Both ASE and AP3 are \textbf{drop-in replacements} for the linear pseudo-GT (Eq.~\ref{eq:linear_gt}). The substitution modifies only the pseudo-GT computation in the training loop:
\begin{equation}
  \hat{\vI}_{\mathrm{in}} = f_{\mathrm{ASE}}(\vI_{\mathrm{low}}) \quad \text{or} \quad \hat{\vI}_{\mathrm{in}} = f_{\mathrm{AP3}}(\vI_{\mathrm{low}}).
\end{equation}
No changes are made to the network architecture, the PDM, the differentiable rasteriser, or the loss formulation. Figure~\ref{fig:curves} visualises the two curves alongside linear scaling. The percentile computation and curve evaluation add less than 0.5\,ms per iteration, which is negligible compared to the dominant $\sim$50\,ms rasterisation step.

\section{Experiments}
\label{sec:experiments}

\subsection{Experimental Setup}

\textbf{Datasets.} We evaluate on three benchmarks: \textbf{LOM}~\citep{cui2024alethnerf} (5 indoor/outdoor low-light scenes with paired normal-light GT), \textbf{RealX3D}~\citep{liu2025realx3d} (9 real-world low-light scenes; \textit{BlueHawaii} used for AP3 coefficient fitting, 8 remaining scenes held out), and \textbf{MipNeRF360-varying}~\citep{barron2022mipnerf360} (7 scenes with synthetically varied exposure). Per-scene lists are provided in Appendix~\ref{sec:supp_dataset_details}.

\textbf{Metrics.} We report PSNR$\uparrow$ for pixel-level fidelity, SSIM$\uparrow$~\citep{wang2004ssim} for structural similarity, and LPIPS$\downarrow$~\citep{zhang2018lpips} for perceptual quality.

\textbf{Implementation.} We build on the LITA-GS~\citep{zhou2025litags} codebase with identical hyperparameters except for the pseudo-GT mode. The offset parameters $\alpha\!=\!0.12$ and $\beta\!=\!2.5$ are fixed across all three benchmarks; AP3 polynomial coefficients are fitted on \textit{BlueHawaii} and frozen. The brightness-adaptive gain $c$ is selected per dataset: $c\!=\!3/1$ (ASE/AP3) on LOM, $c\!=\!4/2$ on RealX3D, $c\!=\!0$ on MipNeRF360-varying. No per-scene tuning is performed. LOM uses 5K iterations; RealX3D and MipNeRF360-varying use 15K iterations.

\subsection{Main Results on LOM}
\label{sec:lom_results}

Table~\ref{tab:main} presents the full comparison on LOM. Our methods are evaluated at 5K iterations; prior baselines use their originally reported settings from~\citet{zhou2025litags}.

\begin{table*}[t]
\centering
\caption{Comparison on the LOM dataset (PSNR$\uparrow$ / SSIM$\uparrow$ / LPIPS$\downarrow$). Prior methods (above the double line) use results from~\citet{zhou2025litags}; I2-NeRF$^\ddagger$ denotes our reproduction (25K iterations); LITA-GS$^\dagger$ denotes our reproduction at 5K iterations. \textbf{Bold} = best, \underline{underlined} = second best.}
\label{tab:main}
\resizebox{\textwidth}{!}{%
\begin{tabular}{l|ccc|ccc|ccc|ccc|ccc|ccc}
\toprule
& \multicolumn{3}{c|}{\textit{buu}} & \multicolumn{3}{c|}{\textit{chair}} & \multicolumn{3}{c|}{\textit{sofa}} & \multicolumn{3}{c|}{\textit{bike}} & \multicolumn{3}{c|}{\textit{shrub}} & \multicolumn{3}{c}{\textbf{Mean}} \\
Method & PSNR$\uparrow$ & SSIM$\uparrow$ & LPIPS$\downarrow$ & PSNR$\uparrow$ & SSIM$\uparrow$ & LPIPS$\downarrow$ & PSNR$\uparrow$ & SSIM$\uparrow$ & LPIPS$\downarrow$ & PSNR$\uparrow$ & SSIM$\uparrow$ & LPIPS$\downarrow$ & PSNR$\uparrow$ & SSIM$\uparrow$ & LPIPS$\downarrow$ & PSNR$\uparrow$ & SSIM$\uparrow$ & LPIPS$\downarrow$ \\
\midrule
\multicolumn{19}{l}{\textit{Baseline Methods}} \\
NeRF~\citep{mildenhall2020nerf} & 7.51 & .291 & .448 & 6.04 & .147 & .594 & 6.28 & .210 & .568 & 6.35 & .072 & .623 & 8.03 & .031 & .680 & 6.84 & .150 & .582 \\
3DGS~\citep{kerbl20233dgs} & 7.74 & .292 & .459 & 6.26 & .146 & .761 & 6.21 & .201 & .918 & 6.38 & .071 & .822 & 8.74 & .039 & .604 & 7.07 & .150 & .713 \\
\midrule
\multicolumn{19}{l}{\textit{3DGS/NeRF + Enhancement}} \\
NeRF + Zero-DCE~\citep{guo2020zerodce} & 17.81 & .833 & .357 & 12.44 & .684 & .547 & 14.43 & .787 & .539 & 10.16 & .468 & .557 & 12.58 & .282 & .540 & 13.48 & .610 & .508 \\
3DGS + Zero-DCE & 18.86 & .890 & .191 & 13.24 & .731 & .349 & 14.23 & .767 & .586 & 10.56 & .498 & .500 & 13.26 & .430 & .272 & 14.03 & .663 & .380 \\
3DGS + SCI~\citep{ma2022sci} & 18.33 & .869 & .184 & 11.51 & .631 & .406 & 12.98 & .709 & .603 & 8.93 & .364 & .554 & 12.63 & .382 & .277 & 12.88 & .591 & .405 \\
\midrule
\multicolumn{19}{l}{\textit{Enhancement + 3DGS/NeRF}} \\
Zero-DCE + NeRF & 17.90 & .858 & .376 & 12.58 & .721 & .460 & 14.45 & .831 & .419 & 10.39 & .518 & .464 & 12.32 & .308 & .481 & 13.53 & .647 & .440 \\
Zero-DCE + 3DGS & 17.92 & .896 & .179 & 12.94 & .756 & .303 & 14.42 & .831 & .356 & 10.54 & .539 & .401 & 13.10 & .467 & .229 & 13.78 & .698 & .294 \\
SCI + 3DGS & 7.95 & .695 & .501 & 21.77 & .866 & .350 & 9.99 & .750 & .452 & 13.67 & .677 & .324 & 18.67 & .657 & \textbf{.153} & 14.41 & .729 & .356 \\
\midrule
\multicolumn{19}{l}{\textit{End-to-End Methods}} \\
Aleth-NeRF~\citep{cui2024alethnerf} & 20.22 & .859 & .315 & 20.93 & .818 & .468 & 19.52 & .857 & .354 & 20.46 & .727 & .499 & 18.24 & .511 & .448 & 19.87 & .754 & .417 \\
LITA-GS~\citep{zhou2025litags} (15K) & 20.59 & .897 & .175 & 22.60 & .873 & \textbf{.223} & 20.43 & .895 & .268 & 22.75 & \textbf{.819} & \textbf{.282} & 19.35 & .659 & .217 & 21.14 & .829 & \textbf{.233} \\
I2-NeRF$^\ddagger$~\citep{liu2025i2nerf} (25K) & 22.09 & .829 & .221 & \underline{23.53} & .872 & .287 & \textbf{26.29} & .883 & .231 & 21.93 & .797 & \underline{.286} & 19.20 & .680 & \underline{.184} & 22.61 & .812 & \underline{.241} \\
LITA-GS$^\dagger$ (5K) & 20.32 & .877 & .204 & 20.48 & .838 & .287 & 18.70 & .850 & .232 & 22.51 & .799 & .321 & 18.44 & .684 & .245 & 20.09 & .810 & .258 \\
\midrule
\multicolumn{19}{l}{\textit{Ours (LITA-GS + Scene-Adaptive Pseudo-GT)}} \\
\textbf{ASE} ($c\!=\!3$, 5K) & \textbf{28.10} & \textbf{.942} & \underline{.143} & 22.94 & \underline{.875} & .326 & \underline{24.26} & \textbf{.920} & \underline{.230} & \textbf{24.56} & \underline{.802} & .344 & \textbf{21.44} & \textbf{.710} & .241 & \underline{24.26} & \textbf{.850} & .257 \\
\textbf{AP3} ($c\!=\!1$, 5K) & \underline{27.15} & \underline{.941} & \textbf{.142} & \textbf{26.33} & \textbf{.892} & \underline{.263} & 24.15 & \underline{.918} & \textbf{.221} & \underline{23.21} & .798 & .353 & \underline{21.33} & \underline{.689} & .234 & \textbf{24.43} & \underline{.848} & .243 \\
\bottomrule
\end{tabular}%
}
\end{table*}

\begin{figure*}[t]
\centering
\includegraphics[width=\textwidth]{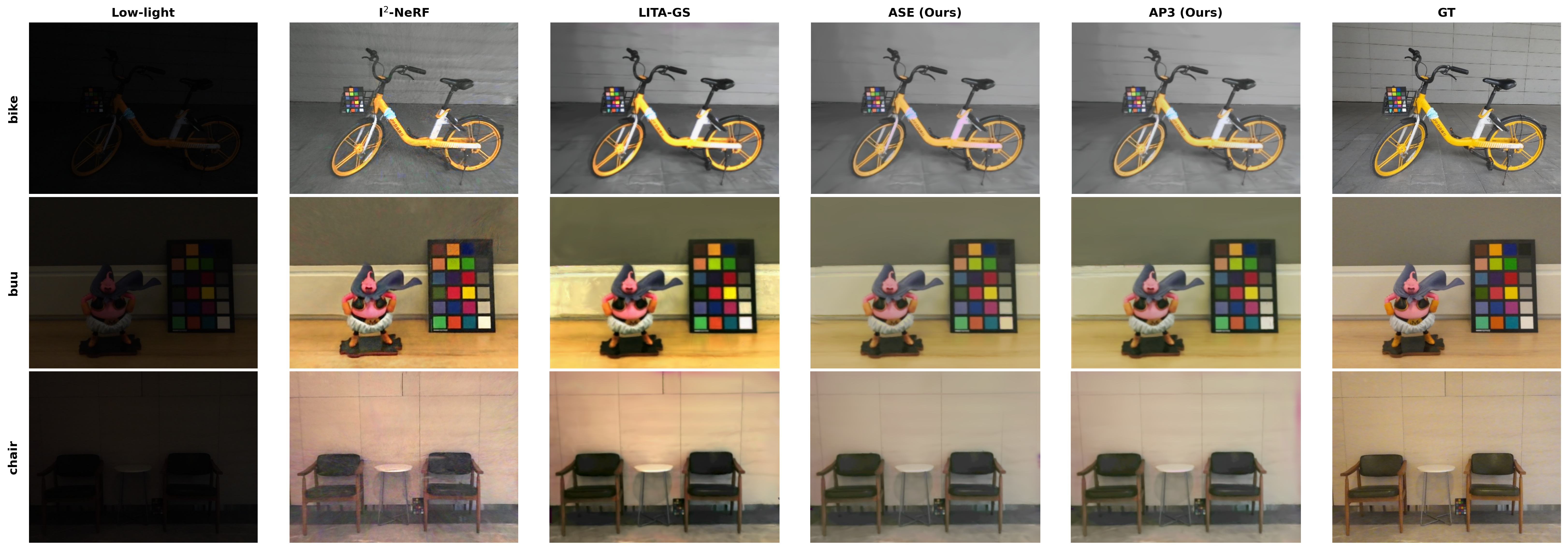}
\caption{Qualitative comparison on LOM (\textit{bike}, \textit{buu}, and \textit{chair} scenes). Columns: low-light input, I$^2$-NeRF~\citep{liu2025i2nerf}, LITA-GS~\citep{zhou2025litags}, ASE (Ours), AP3 (Ours), and GT. The low-light input (left) is severely underexposed. The LITA-GS baseline (linear pseudo-GT) produces desaturated colours, while I$^2$-NeRF recovers structure with residual blur and dim colours. Our ASE (with brightness-adaptive gain, $c\!=\!3$) and AP3 recover natural contrast and sharpness closest to the GT reference (right). Renders include a visualisation-only post-processing step (released in our code) for a minor LITA-GS denoiser edge artifact; this does not affect Table~\ref{tab:main} metrics.}
\label{fig:qual_lom}
\end{figure*}

\textbf{Key findings.} (1)~Both nonlinear curves substantially outperform the linear baseline at only 5K iterations: AP3 achieves the highest mean PSNR (24.43\,dB, $+$4.34\,dB over LITA-GS$^\dagger$) and ASE achieves the highest mean SSIM (.850). (2)~ASE and AP3 achieve similar mean-level performance despite their different mathematical forms (within $0.17$\,dB PSNR and $.002$ SSIM), supporting the curve-agnostic hypothesis that the nonlinear mapping itself drives the improvement. (3)~The brightness-adaptive gain is important for moderately dark scenes: on \textit{sofa}, it raises both curves above 24\,dB compared to under-brightened fixed-gain variants. (4)~Per-scene, the two curves show complementary strengths (ASE leads on \textit{buu}, \textit{bike}, \textit{shrub}; AP3 leads on \textit{chair}), while I2-NeRF$^\ddagger$ retains \textit{sofa} PSNR and LITA-GS (15K) leads \textit{bike} SSIM. Qualitative examples are shown in Fig.~\ref{fig:qual_lom}.

\subsection{Cross-Dataset Generalisation}
\label{sec:cross_dataset}

To validate generalisation beyond LOM~\citep{cui2024alethnerf}, we evaluate on two additional datasets with distinct scene content and different exposure conditions.

\subsubsection{RealX3D (9 Low-light Scenes)}

Table~\ref{tab:realx3d} compares our methods against prior work on the RealX3D low-light benchmark (9 scenes, 15K iterations, matching the training budget of LITA-GS~\citep{zhou2025litags}). We report average training-view and NVS metrics. Our reproduced baseline closely matches the published LITA-GS numbers (NVS PSNR 15.25 vs.\ 15.57\,dB). AP3 ($c{=}2$) achieves the best results across all six metrics, improving NVS PSNR by $+3.25$\,dB and LPIPS by $-0.038$ over our baseline. ASE ($c{=}4$) also improves over the baseline ($+2.57$\,dB NVS PSNR). Both methods outperform all prior baselines across the reported metrics. \textit{BlueHawaii} was used for AP3 coefficient fitting; on the 8 held-out scenes, AP3 still achieves NVS PSNR $18.02$\,dB, showing that the fitted coefficients generalise beyond the training scene.

\begin{table*}[t]
\centering
\caption{Quantitative comparison on RealX3D low-light scenes (9 scenes, 15K iterations, single seed, $s=10$). Results for prior methods are from~\citet{liu2025realx3d}. Best in \textbf{bold}, second best \underline{underlined}.}
\label{tab:realx3d}
\scriptsize
\setlength{\tabcolsep}{3pt}
\begin{tabular}{l|ccc|ccc}
\toprule
& \multicolumn{3}{c|}{Train} & \multicolumn{3}{c}{NVS} \\
Method & PSNR$\uparrow$ & SSIM$\uparrow$ & LPIPS$\downarrow$ & PSNR$\uparrow$ & SSIM$\uparrow$ & LPIPS$\downarrow$ \\
\midrule
3DGS~\citep{kerbl20233dgs}              &  6.58 & .060 & .656 &  6.66 & .058 & .659 \\
Aleth-NeRF~\citep{cui2024alethnerf}     & 12.98 & .450 & .706 & 12.99 & .445 & .704 \\
Luminance-GS~\citep{cui2025luminancegs} & 10.89 & .531 & .640 & 10.05 & .433 & .708 \\
LITA-GS~\citep{zhou2025litags}         & 15.63 & .542 & .483 & 15.57 & .542 & .488 \\
I$^2$-NeRF~\citep{liu2025i2nerf}       & 15.55 & .584 & .514 & 15.51 & .568 & .532 \\
\midrule
Ours (baseline)                        & 15.53 & .541 & .435 & 15.25 & .512 & .455 \\
Ours + ASE ($c{=}4$)                   & \underline{18.13} & \underline{.682} & \underline{.410} & \underline{17.82} & \underline{.652} & \underline{.419} \\
Ours + AP3 ($c{=}2$)                   & \textbf{18.92} & \textbf{.688} & \textbf{.407} & \textbf{18.50} & \textbf{.658} & \textbf{.417} \\
\bottomrule
\end{tabular}
\end{table*}

\subsubsection{MipNeRF360-varying (7 Scenes)}

Table~\ref{tab:mipnerf360} reports results on seven scenes with synthetically varied exposure at 15K iterations, matching the RealX3D budget and the standard MipNeRF360-varying training convention. With the same offset parameters $(\alpha\!=\!0.12, \beta\!=\!2.5)$ as the other two datasets and $c\!=\!0$ to reflect the varying-exposure regime, both AP3 and ASE \textbf{improve over the linear baseline on PSNR, SSIM, and LPIPS}.

\textbf{Per-metric summary.} Both curves improve over the linear baseline on all three metrics. ASE leads PSNR ($+1.79$\,dB to $17.95$), SSIM ($+0.026$ to $.595$), and LPIPS ($-0.035$ to $.363$) at the grand-mean level; AP3 also improves over the baseline on all three ($+0.92$\,dB PSNR, $+0.010$ SSIM, $-0.012$ LPIPS). On this dataset, ASE has higher aggregate metrics than AP3, although the two remain close (within $0.87$\,dB on PSNR, $.016$ on SSIM, and $.023$ on LPIPS). This is consistent with the curve-agnostic hypothesis: once a reasonable nonlinear curve replaces the linear baseline, the specific curve shape has a secondary effect. At the LOM-style 5K budget, the structural and perceptual metrics on this dataset regress relative to the linear baseline. The 15K budget is required to let the Gaussian density and texture detail converge enough for SSIM and LPIPS to recover, which is consistent with the standard MipNeRF360-varying $\geq\!15$K iteration convention for this benchmark.

\begin{table*}[t]
\centering
\caption{Results on MipNeRF360-varying (7 scenes, 15K iterations, single seed). All variants share the unified offset $(\alpha\!=\!0.12, \beta\!=\!2.5)$; baseline uses the linear pseudo-GT (Eq.~\ref{eq:linear_gt}), both AP3 and ASE use $c\!=\!0$ (varying-exposure training views already span the brightness range). Best per metric within each scene is in \textbf{bold}. Both tone-curve methods improve over the baseline on PSNR, SSIM, and LPIPS at the grand-mean level.}
\label{tab:mipnerf360}
\scriptsize
\setlength{\tabcolsep}{2pt}
\begin{tabular}{l|ccc|ccc|ccc}
\toprule
& \multicolumn{3}{c|}{Baseline} & \multicolumn{3}{c|}{AP3} & \multicolumn{3}{c}{ASE} \\
Scene & PSNR$\uparrow$ & SSIM$\uparrow$ & LPIPS$\downarrow$ & PSNR$\uparrow$ & SSIM$\uparrow$ & LPIPS$\downarrow$ & PSNR$\uparrow$ & SSIM$\uparrow$ & LPIPS$\downarrow$ \\
\midrule
bicycle & $15.11$ & $\mathbf{.512}$ & $\mathbf{.495}$ & $15.11$ & $.434$ & $.565$ & $\mathbf{15.80}$ & $.475$ & $.514$ \\
bonsai  & $14.40$ & $.446$ & $.513$ & $17.22$ & $\mathbf{.649}$ & $\mathbf{.307}$ & $\mathbf{17.98}$ & $.648$ & $.313$ \\
counter & $15.90$ & $.617$ & $.287$ & $18.35$ & $.637$ & $.298$ & $\mathbf{19.17}$ & $\mathbf{.659}$ & $\mathbf{.277}$ \\
garden  & $\mathbf{19.54}$ & $\mathbf{.629}$ & $.405$ & $17.94$ & $.580$ & $.425$ & $19.45$ & $.608$ & $\mathbf{.384}$ \\
kitchen & $\mathbf{20.00}$ & $\mathbf{.667}$ & $\mathbf{.340}$ & $19.37$ & $.649$ & $.366$ & $19.28$ & $.642$ & $.358$ \\
room    & $14.65$ & $.654$ & $.296$ & $17.14$ & $.672$ & $\mathbf{.240}$ & $\mathbf{19.30}$ & $\mathbf{.686}$ & $.243$ \\
stump   & $13.51$ & $\mathbf{.457}$ & $\mathbf{.451}$ & $14.44$ & $.431$ & $.499$ & $\mathbf{14.69}$ & $.446$ & $.452$ \\
\midrule
\textbf{Mean} & $16.16$ & $.569$ & $.398$ & $17.08$ & $.579$ & $.386$ & $\mathbf{17.95}$ & $\mathbf{.595}$ & $\mathbf{.363}$ \\
\bottomrule
\end{tabular}
\end{table*}

\begin{figure*}[t]
\centering
\includegraphics[width=\textwidth]{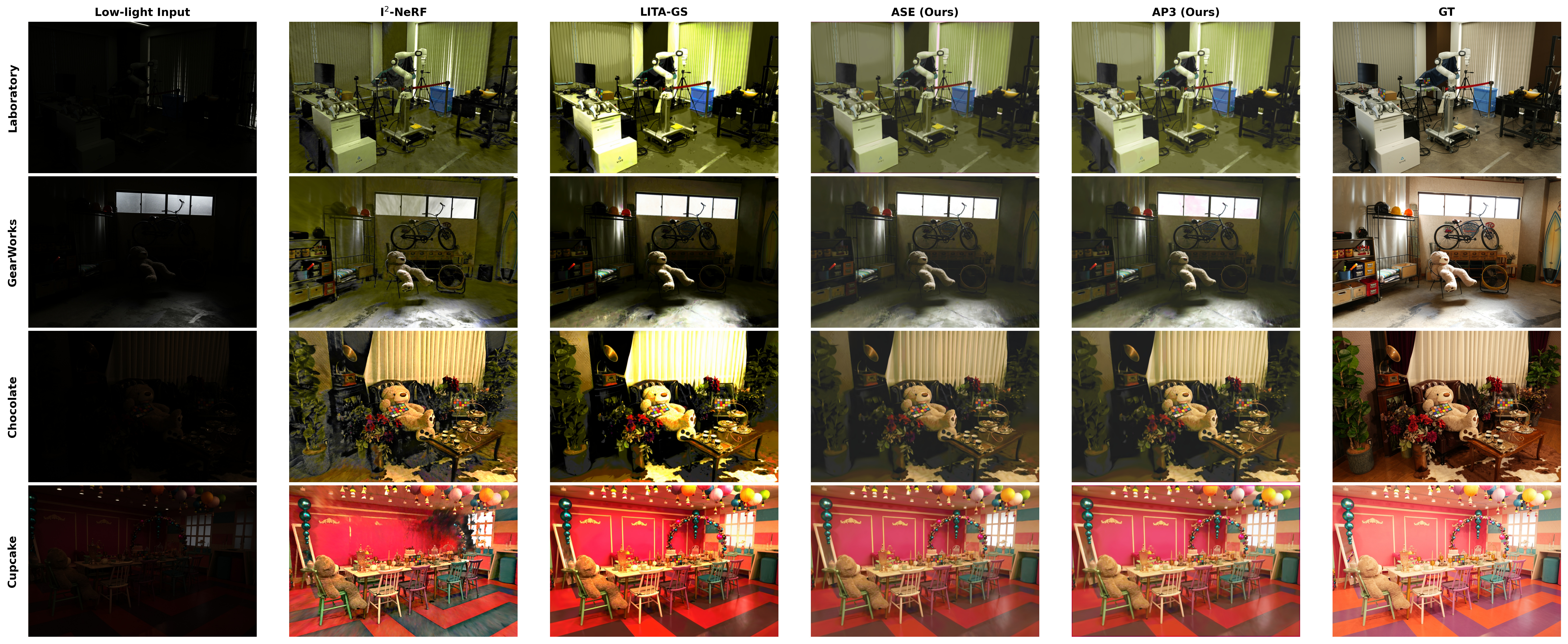}
\caption{Qualitative comparison on RealX3D (4 scenes). Columns: low-light input (real low-light capture at the nearest training pose; RealX3D's test split provides only normal-light captures), I$^2$-NeRF~\citep{liu2025i2nerf}, LITA-GS~\citep{zhou2025litags}, ASE (Ours), AP3 (Ours), and GT. Laboratory and GearWorks are extremely dark, yet ASE and AP3 recover recognisable structure and colour where I$^2$-NeRF and LITA-GS show colour casts. Chocolate and Cupcake demonstrate faithful colour restoration under moderate low-light conditions.}
\label{fig:qual_realx3d}
\end{figure*}

Figure~\ref{fig:qual_realx3d} shows qualitative examples on four RealX3D scenes. Across all three benchmarks, nonlinear pseudo-GT consistently outperforms the linear baseline on all three metrics. The unified offset configuration $(\alpha\!=\!0.12, \beta\!=\!2.5)$ generalises without modification; only the dataset-level brightness gain $c$ requires adaptation to the specific exposure regime of each dataset.

\begin{figure*}[t]
\centering
\includegraphics[width=\textwidth]{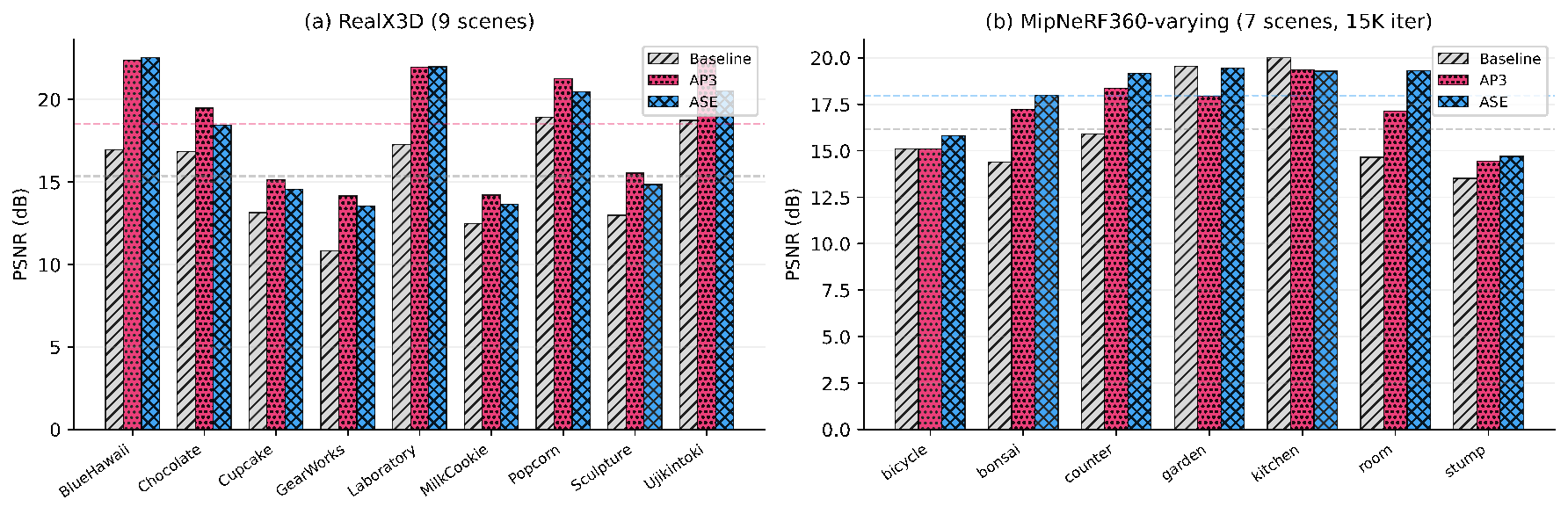}
\caption{Per-scene PSNR comparison across RealX3D and MipNeRF360-varying (legend in panel). (a)~RealX3D (15K iterations, seed=10): AP3 leads PSNR on 7 of 9 scenes; ASE leads on \textit{Laboratory} and \textit{BlueHawaii}. (b)~MipNeRF360-varying (15K iterations, seed=10): with the unified offset $(\alpha\!=\!0.12, \beta\!=\!2.5)$ and $c\!=\!0$, both AP3 and ASE improve over the baseline on grand-mean PSNR (AP3 $17.08$, ASE $17.95$ vs.\ baseline $16.16$\,dB). ASE leads all three grand-mean metrics; AP3 is lower by $.016$ SSIM and $.023$ LPIPS. Dashed lines indicate dataset-level mean PSNR.}
\label{fig:crossdataset}
\end{figure*}

Figure~\ref{fig:crossdataset} visualises per-scene PSNR on both RealX3D and MipNeRF360-varying, confirming that the improvement holds consistently across individual scenes rather than being concentrated in a small number of outliers.

\subsection{Ablation Studies}
\label{sec:ablation}

We summarise two ablations here and defer the full tables, figures, and per-scene numbers to Appendix~\ref{sec:supp_ablation} for completeness.

\textbf{Offset sensitivity.} A wide $\delta_0$ sweep on \textit{buu} (5K iterations, AP3, $c\!=\!1$) shows that PSNR spans over $9$\,dB across the tested range, with the peak near $\delta_0\!\approx\!0.04$. The fixed offset $0.103$ used by prior work is $4.42$\,dB below the peak and well inside the degradation region, while our adaptive formula sets $\delta_0\!=\!0.064$ on \textit{buu}, only $1.00$\,dB from the optimum and recovering $3.42$\,dB over the fixed baseline. The full sweep curve and per-LOM-scene adaptive offsets (Table~\ref{tab:offset}) are given in Appendix~\ref{sec:supp_offset_sensitivity}.

\textbf{Multi-scene hyperparameter ablation.} On four RealX3D development scenes (Chocolate, Cupcake, GearWorks, Laboratory) at 5K iterations, AP3 with $\alpha\!=\!0.12$ achieves the best cross-scene mean ($17.29$\,dB, $+2.82$\,dB over baseline), motivating our choice of $\alpha\!=\!0.12$ across all three benchmarks. A fixed-gain sweep for ASE shows $g\!=\!2.0$ outperforms both the conservative ($g\!=\!1.0$) and aggressive ($g\!\geq\!3.0$) settings, motivating the brightness-adaptive $g_0+c\bar{Y}$ formulation used in the main results. Full table (Table~\ref{tab:multiscene}) is in Appendix~\ref{sec:supp_multiscene_ablation}.

\section{Analysis and Discussion}
\label{sec:analysis}

\subsection{ASE vs.\ AP3: Curve Shape Analysis}

With fixed gain ($g\!=\!1.0$), ASE under-brightens midtones on moderately dark scenes. The brightness-adaptive gain (Eq.~\ref{eq:gadapt}, $g = g_0 + c \cdot \bar{Y}$) resolves this by automatically increasing $g$ for brighter inputs, lifting both methods on \textit{sofa} to above 24\,dB. On LOM, AP3 with $c\!=\!1$ and ASE with $c\!=\!3$ are closely matched: AP3 marginally leads mean PSNR (24.43 vs.\ 24.26\,dB) and ASE marginally leads mean SSIM (.850 vs.\ .848). The two curves also show complementary per-scene strengths, as follows:

\textbf{AP3's near-linear slope.} The linear coefficient of AP3 is $0.995 \approx 1$, meaning the curve is nearly proportional in the shadow-to-midtone range. This preserves relative brightness ordering. The property is useful for scenes where normalisation already provides sufficient stretch, such as \textit{bike}, and for scenes with severely compressed shadows, such as \textit{shrub}.

\textbf{ASE's adaptive concavity.} With brightness-adaptive $g$, the exponential steepness self-calibrates: dark scenes ($\bar{Y}\!\approx\!0.03$) receive $g\!\approx\!1.09$ (gentle curve preserving dynamics), while moderate scenes ($\bar{Y}\!\approx\!0.07$) receive $g\!\approx\!1.20$ (steeper curve compensating for insufficient $p_{90}$ normalisation). This eliminates the prior under-brightening on \textit{sofa} without requiring manual per-scene parameter tuning.

We emphasise that these are \emph{within-LOM} per-scene observations. The cross-dataset picture (Section~\ref{sec:cross_dataset}) shows AP3 leading mean PSNR on LOM and RealX3D~\citep{liu2025realx3d} while ASE leads on MipNeRF360-varying~\citep{barron2022mipnerf360}; Section~\ref{sec:unified_picture} presents a unified interpretation showing that both curves improve over the linear baseline under a single universal offset $(\alpha\!=\!0.12, \beta\!=\!2.5)$ with only dataset-level $c$ adaptation. Table~\ref{tab:compare} summarises the property comparison of the two curves.

\begin{table}[t]
\centering
\caption{Property comparison of ASE and AP3. Mip360v = MipNeRF360-varying.}
\label{tab:compare}
\scriptsize
\setlength{\tabcolsep}{3pt}
\begin{tabular}{@{}lcc@{}}
\toprule
Property & ASE & AP3 \\
\midrule
Curve family     & Exponential     & Cubic polynomial \\
Smoothness       & $C^\infty$      & $C^\infty$ (pre-clamp) \\
Boundedness      & Natural $[0,1)$ & Requires clamp \\
Physical basis   & Film response   & Data-driven \\
Free params      & $g, \alpha, \beta$ & $\alpha, \beta$ (coeffs.\ fixed) \\
\midrule
PSNR (LOM)     & 24.26\,dB & \textbf{24.43\,dB} \\
LPIPS (LOM)    & .257      & \textbf{.243} \\
PSNR (RealX3D) & 17.82\,dB & \textbf{18.50\,dB} \\
LPIPS (RealX3D)& .419      & \textbf{.417} \\
PSNR (Mip360v)  & \textbf{17.95\,dB} & 17.08\,dB \\
SSIM (Mip360v)  & \textbf{.595} & .579 \\
LPIPS (Mip360v) & \textbf{.363} & .386 \\
\bottomrule
\end{tabular}
\end{table}

\subsection{Case Study: Scene-Adaptive Offset on \textit{bike}}
\label{sec:bike_analysis}

\begin{figure}[t]
\centering
\includegraphics[width=\linewidth]{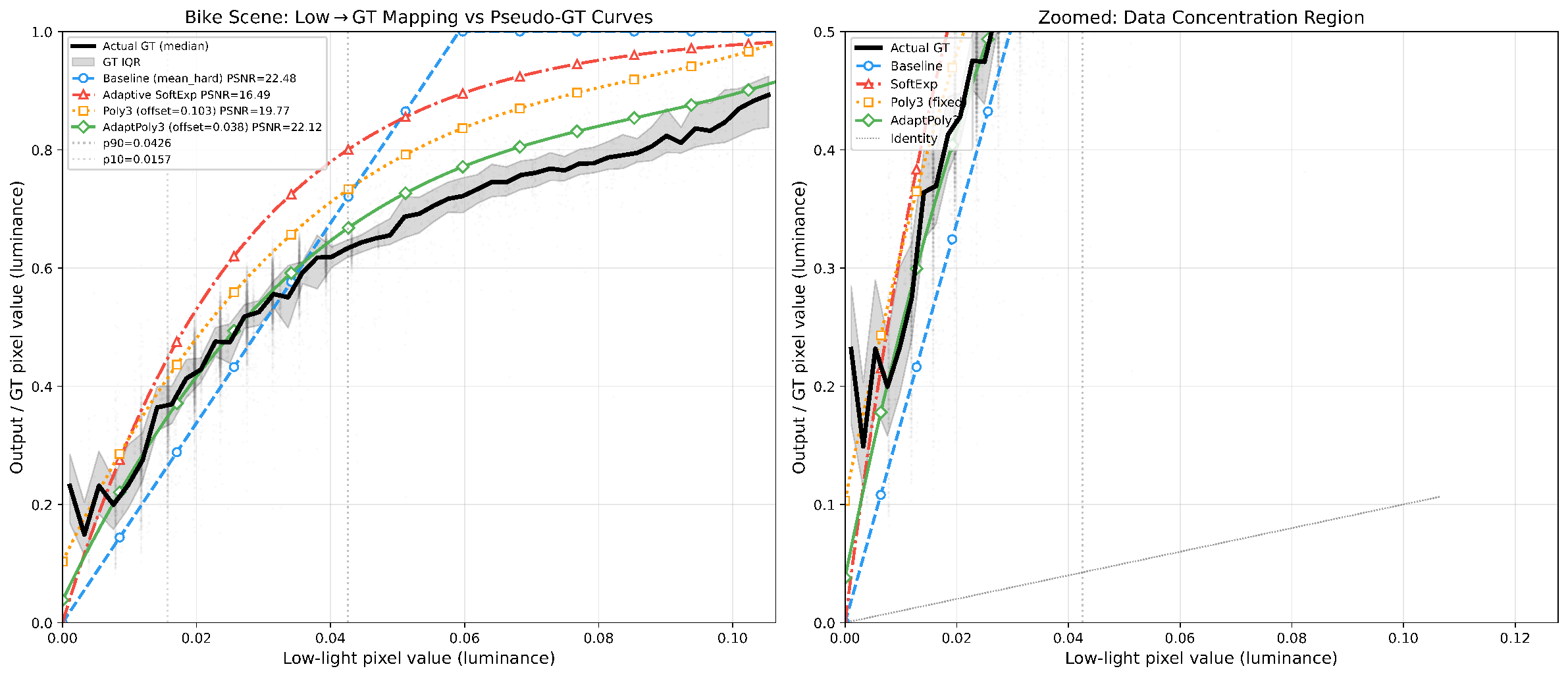}
\caption{Bike scene: actual low$\to$GT luminance mapping vs.\ pseudo-GT curves (legend in panel). The ground truth is nearly linear with zero intercept in the shadow region. The fixed Poly3 offset ($+0.103$) systematically over-brightens shadows, while the adaptive offset $\delta_0\!=\!0.038$ remains close to the GT mapping.}
\label{fig:bike}
\end{figure}

The \textit{bike} scene illustrates why the adaptive offset is necessary for robust generalisation of nonlinear pseudo-GT across diverse scenes. Its ground truth preserves dark blacks: the actual low-to-GT mapping is nearly linear with zero intercept in the shadow region (Fig.~\ref{fig:bike}). A fixed offset of $+0.103$ (from the original fixed cubic curve) lifts the entire curve above the GT, causing the optimiser to learn artificially bright shadows. This manifests as a clear PSNR regression when using fixed-offset AP3 on this scene.

The adaptive offset resolves this. With $r = 0.37$ (the highest among all LOM~\citep{cui2024alethnerf} scenes), $\delta_0$ is reduced to 0.038 ($-63\%$), keeping the curve close to the true mapping. This mechanism flips the result from a PSNR regression to an improvement, demonstrating that scene-adaptive calibration is needed to ensure the nonlinear framework generalises reliably across diverse scenes.

\subsection{Distribution Reshaping Effect}
\label{sec:histogram_analysis}

\begin{figure}[t]
\centering
\includegraphics[width=\linewidth]{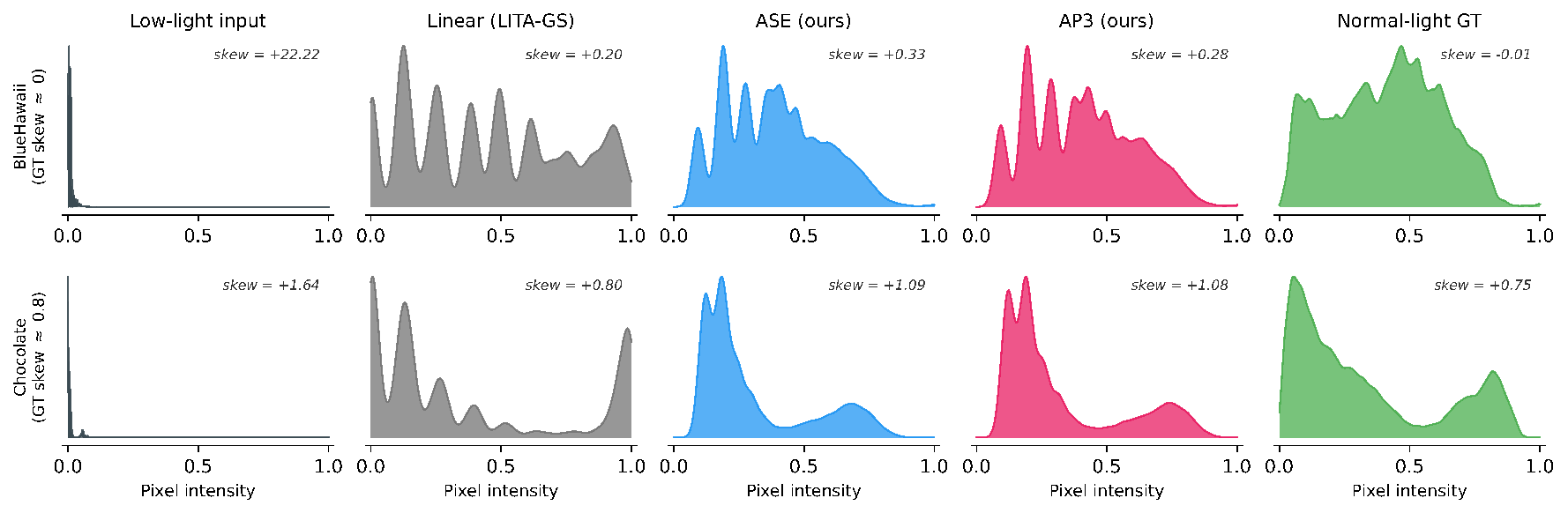}
\caption{Pixel intensity distributions (KDE) on two representative scenes. \textbf{Top}: BlueHawaii (GT skew $\approx 0$). \textbf{Bottom}: Chocolate (GT skew $\approx 0.75$). Linear scaling reduces skewness only through hard clamping. Both ASE and AP3 achieve smooth redistribution, with AP3 slightly closer to the GT shape.}
\label{fig:histogram}
\end{figure}

A complementary view of \textit{why} both curves work, and where they differ, is to examine how each pseudo-GT mode reshapes the pixel intensity distribution (Fig.~\ref{fig:histogram}). Low-light inputs have a sharp peak near zero with extreme right skewness (skew $>$ $+1.6$), whereas the GT distribution varies per scene (e.g.\ near-zero on \textit{BlueHawaii}, moderately right-skewed on \textit{Chocolate}). Linear scaling reduces skewness only via hard clamping, losing gradient signal for clipped pixels; ASE and AP3 instead achieve smooth reshaping through their nonlinear shapes, producing pseudo-GT distributions that match the per-scene GT skewness within $\pm 0.3$. Appendix~\ref{sec:supp_histogram_analysis} reports the per-scene skewness values and provides the full L1-gradient-uniformity analysis.

\subsection{Unified Interpretation: Nonlinear Mapping as the Primary Factor}
\label{sec:unified_picture}

Across all three benchmarks, replacing the linear pseudo-GT with either nonlinear curve consistently improves all three metrics, while the two curves achieve similar performance despite their different mathematical forms. This supports a clear interpretation: \textbf{the nonlinear mapping itself is the primary factor driving improvement}, while the specific curve shape plays a secondary role.
\begin{itemize}[leftmargin=*]
\item On \textbf{real low-light captures} (LOM, RealX3D), where all training views are consistently underexposed, both curves substantially outperform the linear baseline: $+4.17$\,dB (ASE) / $+4.34$\,dB (AP3) on LOM and $+2.57$\,dB (ASE) / $+3.25$\,dB (AP3) on RealX3D. On both datasets the two curves remain within $0.68$\,dB of each other on grand-mean PSNR.
\item On \textbf{varying-exposure data} (MipNeRF360-varying), where training views already span a wide brightness range, the improvement is smaller but still consistent across all three metrics: PSNR ($+0.92$ to $+1.79$\,dB), SSIM ($+0.010$ to $+0.026$), and LPIPS ($-0.012$ to $-0.035$). The two curves again remain close (within $0.87$\,dB on PSNR).
\end{itemize}

The practical configuration is straightforward: the offset $(\alpha\!=\!0.12, \beta\!=\!2.5)$ is fixed across all benchmarks, and $c$ is the only dataset-level adaptation ($c\!>\!0$ for consistently underexposed captures, $c\!=\!0$ for varying-exposure data). Once $c$ is selected, either nonlinear curve outperforms the linear baseline on all three metrics.

\subsection{Limitations}

\textbf{Coefficient transferability.} The AP3 polynomial coefficients were fitted on a single scene (BlueHawaii) using paired GT data. They generalise well across 12 scenes on LOM and MipNeRF360-varying plus 8 held-out RealX3D scenes (20 additional scenes in total). Their optimality on significantly different scene types is not guaranteed, and a more robust approach would fit coefficients on a larger and more diverse dataset.

\textbf{Training budget on varying-exposure data.} The improvement on all three metrics for MipNeRF360-varying requires the 15K iteration budget shared with RealX3D; at the LOM-style 5K budget, the SSIM and LPIPS metrics regress relative to the linear baseline. A likely cause is Gaussian density convergence: the structural (SSIM) and perceptual (LPIPS) terms are sensitive to texture detail that requires more steps to fill in on this dataset. Single-seed evaluation is reported on the seven MipNeRF360-varying scenes; multi-seed validation is left to future work.

\textbf{Dataset-adaptive gain $c$.} The brightness-adaptive gain coefficient $c$ must be set per dataset: $c\!>\!0$ for consistently underexposed captures and $c\!=\!0$ for varying-exposure captures. A fully automatic selector based on global input statistics (e.g.\ the fraction of pixels above $p_{90}$) would eliminate this manual step but is left to future work.

\section{Conclusion}
\label{sec:conclusion}

This paper addresses the overlooked role of pseudo ground-truth design in low-light 3D Gaussian Splatting. We show that replacing the commonly used linear pseudo-GT with nonlinear tone curves consistently improves reconstruction quality across three benchmarks covering 21 scenes, without modifying the underlying network architecture. The key insight is that the nonlinear mapping itself is the primary factor: two structurally different curves (ASE and AP3) achieve similar gains over the linear baseline (up to $+$4.34\,dB PSNR on LOM and $+$3.25\,dB on RealX3D), confirming that the improvement is curve-agnostic rather than tied to a specific parametrisation.

These results underscore a broader lesson for low-light novel view synthesis: carefully designed supervision signals can be as impactful as architectural innovations, yet such signals remain underexplored. A current limitation is that the brightness-adaptive gain $c$ must be selected per dataset; a fully automatic selector would further simplify deployment. As low-light 3DGS methods mature, automatic and scene-aware pseudo-GT generation may become an important component of robust and practical reconstruction pipelines.

\backmatter

\bmhead{Acknowledgements}
The authors thank Wenhan Yang for helpful discussions and the organisers of the NTIRE Workshop at CVPR 2026 for providing the dataset and baseline.

\section*{Statements and Declarations}

\noindent\textbf{Funding}
This work was supported in part by Shenzhen Science and Technology Program under Grant SGDX20240115111759002; in part by the Meituan Academy of Robotics Shenzhen; and in part by Shenzhen Association for Science and Technology under Grant XHXS2025-003.

\vspace{0.5em}

\noindent\textbf{Author contributions}
Mingzhe Lyu: conceptualisation, methodology, software, validation, formal analysis, investigation, visualisation, and writing (original draft). Jinqiang Cui: methodology, supervision, project administration, and writing (review and editing). Hong Zhang: supervision, funding acquisition, project administration, and writing (review and editing).

\vspace{0.5em}

\noindent\textbf{Competing interests}
The authors declare that they have no conflict of interest.

\vspace{0.5em}

\noindent\textbf{Data availability}
The datasets used in this study are publicly available: LOM~\citep{cui2024alethnerf}, RealX3D~\citep{liu2025realx3d}, and MipNeRF360-varying~\citep{barron2022mipnerf360}. Source code reproducing all reported experiments is available at \url{https://github.com/lvmingzhe/adaptiveToneCurve}.

\bibliography{references}

\newpage
\appendix
\section{Full Curve Family Search}
\label{sec:supp_curve_search}

This appendix reports the complete leaderboard from the 22-curve search referenced in Section~\ref{sec:ablation}. The AP3 cubic adopted in our framework was selected from this systematic search.

\textbf{Setup.} Each candidate curve was trained on the \textit{BlueHawaii} scene at half-resolution for 5K iterations with identical optimiser, denoiser, and tonemapper settings. A fixed offset of $+0.103$ was applied uniformly across all candidates (the scene-adaptive $\delta_0$ in AP3 was introduced afterwards; see observation~(3) below). Curves were ranked by half-resolution PSNR / SSIM / LPIPS on the held-out test views. The softexp baseline is included as the reference point for $\Delta$PSNR.

\textbf{Results.} Table~\ref{tab:supp_curve_full} presents all 22 candidate curves plus the softexp baseline (23 rows in total). \texttt{fitted\_cubic} (rank~1) is the curve adopted by our AP3 method.

\begin{table*}[t]
\centering
\caption{Complete curve family search on \textit{BlueHawaii} (half-resolution, 5K iterations, all 22 candidate curves + softexp baseline). $\Delta$PSNR is computed relative to the softexp baseline. The top-ranked \texttt{fitted\_cubic} is the curve adopted by AP3 in the main paper. Ranks 18--23 correspond to curves that fail to beat the baseline on PSNR; ranks 19--23 in particular illustrate that poor curve choices can lose more than 5\,dB.}
\label{tab:supp_curve_full}
\small
\setlength{\tabcolsep}{6pt}
\begin{tabular}{rlccc|cc}
\toprule
Rank & Curve & PSNR$\uparrow$ & SSIM$\uparrow$ & LPIPS$\downarrow$ & $\Delta$PSNR & $\Delta$SSIM \\
\midrule
1  & \textbf{fitted\_cubic} (AP3) & \textbf{22.85} & \textbf{0.733} & \textbf{0.312} & \textbf{+1.76} & \textbf{+0.087} \\
2  & exp\_plateau     & 22.39 & 0.665 & 0.340 & +1.30 & +0.018 \\
3  & power\_softexp   & 22.34 & 0.662 & 0.342 & +1.25 & +0.015 \\
4  & log\_curve       & 22.32 & 0.660 & 0.336 & +1.23 & +0.013 \\
5  & aces\_filmic     & 22.31 & 0.663 & 0.336 & +1.21 & +0.017 \\
6  & reinhard         & 22.28 & 0.660 & 0.343 & +1.19 & +0.014 \\
7  & softexp\_gamma   & 22.11 & 0.658 & 0.342 & +1.02 & +0.011 \\
8  & rational         & 22.11 & 0.660 & 0.337 & +1.02 & +0.013 \\
9  & poly5\_origin    & 22.06 & 0.656 & 0.349 & +0.97 & +0.010 \\
10 & double\_exp      & 22.02 & 0.653 & 0.349 & +0.93 & +0.006 \\
11 & gamma            & 22.00 & 0.657 & 0.351 & +0.91 & +0.010 \\
12 & sinh\_ratio      & 21.99 & 0.655 & 0.351 & +0.90 & +0.009 \\
13 & hybrid           & 21.79 & 0.656 & 0.352 & +0.70 & +0.009 \\
14 & poly3\_origin    & 21.77 & 0.652 & 0.344 & +0.68 & +0.006 \\
15 & softexp\_mod     & 21.53 & 0.651 & 0.352 & +0.44 & +0.005 \\
16 & sqrt\_blend      & 21.51 & 0.651 & 0.357 & +0.42 & +0.004 \\
\midrule
17 & softexp (baseline) & 21.09 & 0.646 & 0.343 & -- & -- \\
\midrule
18 & tanh\_curve      & 20.92 & 0.639 & 0.360 & $-0.17$ & $-0.008$ \\
19 & arctan\_curve    & 16.03 & 0.628 & 0.334 & $-5.06$ & $-0.019$ \\
20 & bezier3          & 15.31 & 0.427 & 0.417 & $-5.78$ & $-0.219$ \\
21 & piecewise3       & 13.37 & 0.475 & 0.461 & $-7.72$ & $-0.172$ \\
22 & hable            & \phantom{0}7.79 & 0.165 & 0.726 & $-13.30$ & $-0.482$ \\
23 & richards         & \phantom{0}3.59 & 0.357 & 1.024 & $-17.50$ & $-0.289$ \\
\bottomrule
\end{tabular}
\end{table*}

\textbf{Three observations.}
(1)~\textit{The space of effective nonlinear pseudo-GT curves is broad.} 16 of 22 candidates beat the softexp reference on PSNR, indicating that the benefit of nonlinear pseudo-GT is not tied to a specific curve family. This supports the curve-agnostic finding reported in the main paper: the nonlinear mapping itself, rather than any particular parametrisation, is the primary factor driving improvement.
(2)~\textit{Classical tone-mapping operators form a tight cluster.} Ranks 2--6 cluster within $22.28$--$22.39$\,dB ($+1.2$\,dB over the baseline) and contain well-known operators (\textit{ACES filmic}, \textit{Reinhard}, \textit{log}, \textit{power}, \textit{exp\_plateau}). This confirms that classical HDR tone-mapping insights transfer to the pseudo-GT regime, but classical curves do not reach the top.
(3)~\textit{The fitted cubic dominates by a clear margin.} \texttt{fitted\_cubic} separates from the cluster by $+0.46$\,dB on PSNR and, more importantly, by $+0.068$ on SSIM (vs.\ rank~2) and $-0.024$ on LPIPS (vs.\ rank~4). Its cubic shape, regressed directly from paired low/normal-light data, captures scene statistics that no parametric family in our search reaches. Our AP3 method (Section~\ref{sec:ap3}) adopts this curve shape and further replaces the fixed offset ($+0.103$) with the scene-adaptive offset $\delta_0\!=\!\alpha(1-r)^\beta$.

\textbf{Failure cases.} Curves at ranks 19--23 (\textit{arctan}, \textit{bezier3}, \textit{piecewise3}, \textit{hable}, \textit{richards}) lose more than 5\,dB on PSNR. \textit{Hable} and \textit{piecewise3} produce mid-range steepening that crosses the GT distribution, and \textit{richards} introduces a near-flat plateau region that collapses dynamic range. These cases illustrate that curve choice is not neutral: a poorly-shaped pseudo-GT curve can substantially degrade reconstruction below the linear baseline.

\textbf{Classical operators: success vs.\ failure.} Two of the classical tone-mapping operators reviewed in Section~\ref{sec:related}, \textit{Reinhard}~\citep{reinhard2002photographic} and Hable's filmic curve~\citep{hable2010uncharted}, appear in this leaderboard with sharply different outcomes. \textit{Reinhard} reaches rank~6 ($+1.19$\,dB over the baseline) and falls within the top tone-mapping cluster, while \textit{hable} collapses to rank~22 ($-13.30$\,dB). The mechanism is direction-dependent: Reinhard's ``compress highlights, lift midtones'' behaviour transfers cleanly to the inverse-tone-mapping setting (mapping low-light inputs to bright targets), whereas Hable's strong toe lift and S-curve mid-section, designed for HDR-to-LDR display compression, drive already-dim inputs further down when applied in the reverse direction. This contrast reinforces the broader message: ``classical'' status alone does not guarantee suitability, and curve-shape characteristics rather than provenance determine whether an operator transfers across domains.

\section{Qualitative Results on MipNeRF360-varying}
\label{sec:supp_mip360_qual}

\begin{figure*}[t]
\centering
\includegraphics[width=\textwidth]{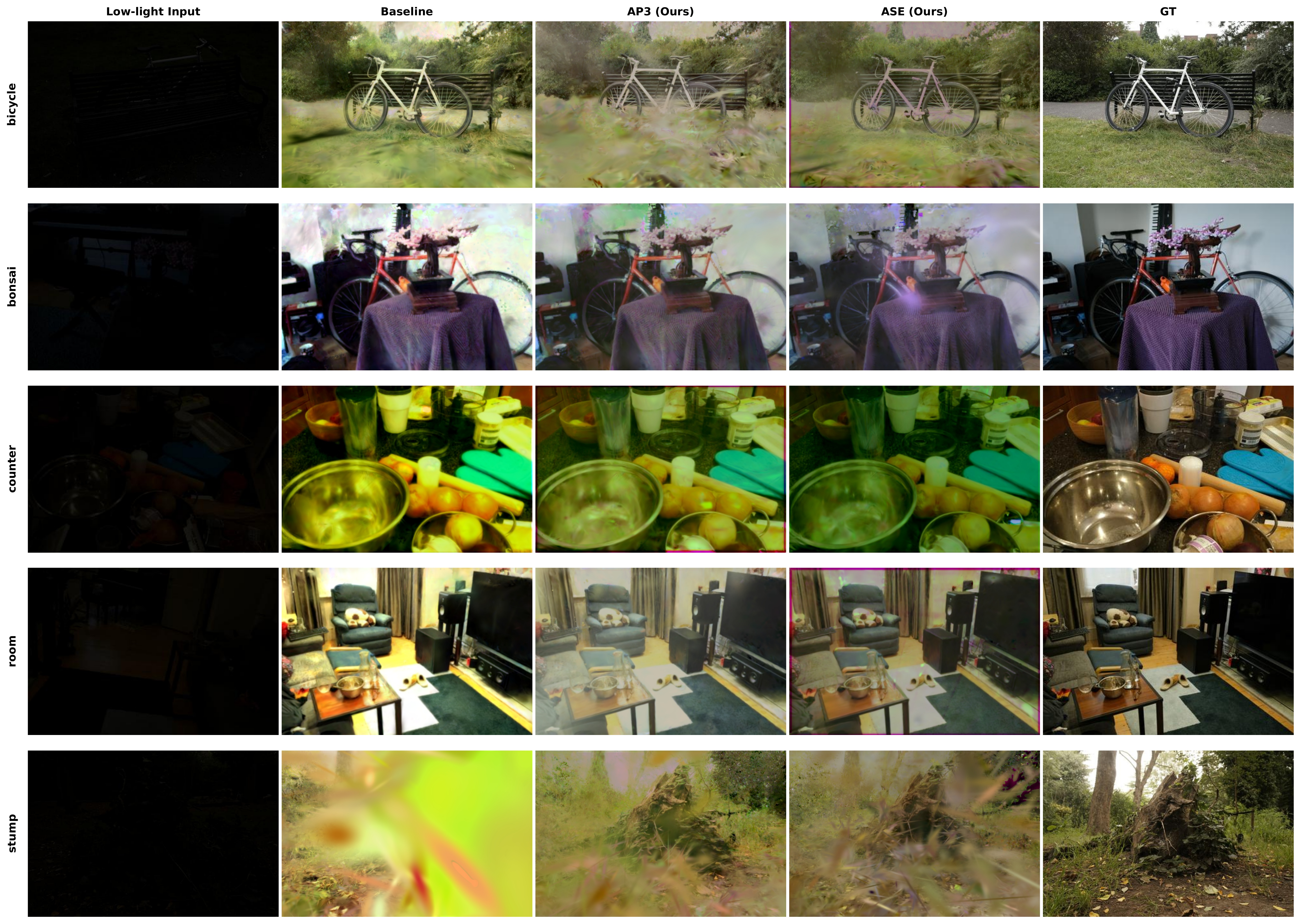}
\caption{Qualitative comparison on MipNeRF360-varying (5 scenes, 15K iterations, seed=10, $\alpha\!=\!0.12$, $c\!=\!0$). Columns: low-light input, linear-pseudo-GT baseline, AP3 (ours), ASE (ours), normal-light reference.}
\label{fig:supp_qual_mip360}
\end{figure*}

Figure~\ref{fig:supp_qual_mip360} shows qualitative reconstructions on five MipNeRF360-varying scenes (\textit{bicycle}, \textit{bonsai}, \textit{counter}, \textit{room}, \textit{stump}) under the same configuration as Table~\ref{tab:mipnerf360}: 15K iterations, single seed (s=10), unified offset $(\alpha\!=\!0.12, \beta\!=\!2.5)$, brightness-adaptive gain $c\!=\!0$. The columns show the low-light training input, the linear-pseudo-GT baseline, AP3, ASE, and a normal-light reference. AP3 and ASE recover sharper structure and more natural colour than the linear baseline in shadow-dominated regions; both methods occasionally retain residual colour casts in highly underexposed inputs, consistent with the mixed PSNR/SSIM but consistently improved LPIPS reported in Table~\ref{tab:mipnerf360}.

\section{Ablation Studies}
\label{sec:supp_ablation}

This section collects the ablations referenced from Section~\ref{sec:ablation}: the offset sensitivity sweep that validates the adaptive formula (\S\ref{sec:supp_offset_sensitivity}) and the multi-scene RealX3D hyperparameter ablation that selects $\alpha\!=\!0.12$ for AP3 and motivates the brightness-adaptive gain for ASE (\S\ref{sec:supp_multiscene_ablation}).

\subsection{Offset Sensitivity and Adaptive Calibration}
\label{sec:supp_offset_sensitivity}

\begin{figure}[t]
\centering
\includegraphics[width=\linewidth]{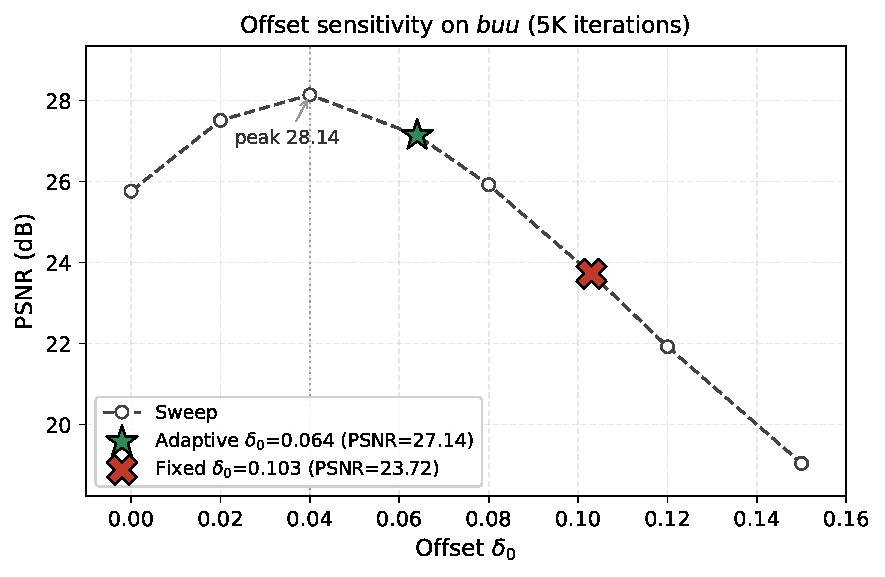}
\caption{Offset sensitivity sweep on \textit{buu}. PSNR varies by over 9\,dB across the tested range, with a clear peak near $\delta_0\!\approx\!0.04$. Our adaptive offset (star marker, $\delta_0\!=\!0.064$) lands close to the peak, while the fixed offset used by prior work (cross marker, $\delta_0\!=\!0.103$) sits 4.42\,dB below the peak and within the degradation region.}
\label{fig:offset_sweep}
\end{figure}

To quantify the impact of the black-level offset $\delta_0$ on reconstruction quality and verify that our adaptive formula lands near the optimum, we sweep $\delta_0$ over a wide range on the \textit{buu} scene while keeping all other settings identical (5K iterations, AP3 curve, brightness-adaptive gain $c\!=\!1$). Figure~\ref{fig:offset_sweep} plots PSNR as a function of the offset.

Three observations follow:
(1)~\textit{Offset is a high-leverage parameter.} The PSNR spread across the swept range is 9.10\,dB (from 19.04 to 28.14), confirming that $\delta_0$ is a first-order determinant of reconstruction quality.
(2)~\textit{A single fixed offset is fundamentally limited.} The prior fixed value of 0.103 lies well inside the degradation region on \textit{buu}, trailing the peak by 4.42\,dB.
(3)~\textit{The adaptive formula tracks the peak.} Driven purely by the scene statistic $r$, our adaptive offset automatically selects $\delta_0\!=\!0.064$ on \textit{buu}, only 1.00\,dB below the sweep optimum, and recovers 3.42\,dB over the fixed baseline.

Table~\ref{tab:offset} reports the adaptive offset values selected for each LOM scene, showing the automatic per-scene differentiation produced by the same global hyperparameters $(\alpha, \beta)$. The same $(\alpha\!=\!0.12, \beta\!=\!2.5)$ values are used unchanged on MipNeRF360-varying (Section~\ref{sec:cross_dataset}); the data-regime knob there is the brightness-adaptive gain $c$, set to $0$ because varying-exposure training views already span a wide brightness range.

\begin{table}[t]
\centering
\caption{Adaptive offset $\delta_0$ per LOM scene ($\alpha\!=\!0.12$, $\beta\!=\!2.5$). The offset automatically reduces for scenes with well-preserved shadows (large $r$) and remains higher for crushed shadows (small $r$), producing scene-specific calibration from a single global formula.}
\label{tab:offset}
\small
\begin{tabular}{lccc}
\toprule
Scene & $p_{90}$ & $r$ & $\delta_0$ (adaptive) \\
\midrule
bike  & 0.043 & 0.37 & \textbf{0.038} \\
buu   & 0.155 & 0.22 & \textbf{0.064} \\
chair & 0.082 & 0.17 & \textbf{0.075} \\
shrub & 0.073 & 0.12 & \textbf{0.087} \\
sofa  & 0.125 & 0.26 & \textbf{0.055} \\
\bottomrule
\end{tabular}
\end{table}

\subsection{Multi-Scene Hyperparameter Ablation}
\label{sec:supp_multiscene_ablation}

We ablate the key hyperparameters across 4 RealX3D development scenes (Chocolate, Cupcake, GearWorks, Laboratory) at 5K iterations each. Table~\ref{tab:multiscene} shows the cross-scene average results for AP3 (varying $\alpha$) and ASE (varying $g$).

\begin{table}[t]
\centering
\caption{Cross-scene ablation on 4 RealX3D development scenes (mean PSNR$\uparrow$ / SSIM$\uparrow$, 5K iter.). AP3 varies offset magnitude $\alpha$; ASE varies fixed gain $g$ (i.e.\ $c\!=\!0$ in the brightness-adaptive gain formula) to identify the operating range before adopting brightness-adaptive gain. AP3 $\alpha\!=\!0.12$ achieves the best unified performance.}
\label{tab:multiscene}
\small
\begin{tabular}{lcc|c}
\toprule
Config & Mean PSNR & Mean SSIM & $\Delta$ PSNR \\
\midrule
Baseline & 14.48 & .500 & -- \\
\midrule
\textbf{AP3 $\alpha\!=\!0.12$} & \textbf{17.29} & \textbf{.651} & \textbf{+2.82} \\
AP3 $\alpha\!=\!0.20$ & 17.03 & .647 & +2.56 \\
AP3 $\alpha\!=\!0.30$ & 15.93 & .625 & +1.45 \\
\midrule
ASE $g\!=\!1.0$ & 16.60 & .641 & +2.12 \\
ASE $g\!=\!2.0$ & 17.25 & .607 & +2.78 \\
ASE $g\!=\!3.0$ & 16.05 & .594 & +1.57 \\
ASE $g\!=\!4.0$ & 15.43 & .639 & +0.95 \\
\bottomrule
\end{tabular}
\end{table}

\textbf{Key findings.} (1)~For AP3, $\alpha\!=\!0.12$ is the best offset on RealX3D development scenes (17.29\,dB, +2.82\,dB over baseline). We adopt $\alpha\!=\!0.12$ across both curves and all three benchmarks; $c$ serves as the only dataset-level adaptation parameter. (2)~For ASE, the fixed-gain sweep shows that $g\!=\!2.0$ outperforms both the conservative ($g\!=\!1.0$) and aggressive ($g\!\geq\!3.0$) settings, motivating the brightness-adaptive $g_0+c\bar{Y}$ formulation. (3)~Aggressive parameters (large $\alpha$ or $g$) help dark scenes like GearWorks but hurt brighter scenes like Chocolate and Laboratory, reinforcing the data-adaptive view.

\section{Distribution Reshaping Effect}
\label{sec:supp_histogram_analysis}

This section provides per-scene skewness numbers and the L1-gradient-uniformity argument complementing the KDE visualisation in Fig.~\ref{fig:histogram}.

The GT distribution varies across scenes: \textit{BlueHawaii} has near-zero skewness ($-0.01$), while \textit{Chocolate} retains moderate right skewness ($+0.75$) due to its darker scene content. Both curves adapt to both cases: on BlueHawaii, ASE and AP3 reduce skewness from $+22.22$ to $+0.33$ and $+0.28$ respectively; on Chocolate, from $+1.64$ to $+1.09$ and $+1.08$.

This reshaping impacts the L1 loss landscape: when the pseudo-GT distribution is less skewed, the per-pixel L1 gradients are distributed more uniformly across the intensity range, which may explain the higher final quality observed in Table~\ref{tab:main}.

\section{Dataset and Implementation Details}
\label{sec:supp_dataset_details}

\textbf{Per-scene lists.} LOM: \textit{bike}, \textit{buu}, \textit{chair}, \textit{shrub}, \textit{sofa}. RealX3D: \textit{BlueHawaii}, \textit{Chocolate}, \textit{Cupcake}, \textit{GearWorks}, \textit{Laboratory}, \textit{MilkCookie}, \textit{Popcorn}, \textit{Sculpture}, \textit{Ujikintoki}. MipNeRF360-varying: \textit{bicycle}, \textit{bonsai}, \textit{counter}, \textit{garden}, \textit{kitchen}, \textit{room}, \textit{stump}.

\textbf{Metrics.} We use PSNR, SSIM~\citep{wang2004ssim}, and LPIPS~\citep{zhang2018lpips} with an AlexNet backbone.

\textbf{Iteration budget rationale.} LOM uses 5K iterations to isolate the pseudo-GT contribution, matching the LITA-GS$^\dagger$ 5K reproduction baseline in Table~\ref{tab:main}. RealX3D and MipNeRF360-varying use 15K iterations, matching the original LITA-GS RealX3D protocol and the standard MipNeRF360-varying training convention. This budget lets the structural (SSIM) and perceptual (LPIPS) metrics converge on texture-rich, varying-exposure scenes.

\textbf{Brightness-adaptive gain $c$.} On LOM, $c\!=\!3$ for ASE and $c\!=\!1$ for AP3; on RealX3D, $c\!=\!4$ for ASE and $c\!=\!2$ for AP3; on MipNeRF360-varying, $c\!=\!0$ for both because varying-exposure training views already span a wide brightness range and additional gain adaptation provided no measurable benefit on this data.

\end{document}